\documentclass{article}

\usepackage{PRIMEarxiv}

\usepackage[utf8]{inputenc} 
\usepackage[T1]{fontenc}    
\usepackage{hyperref}       
\usepackage{url}            
\usepackage{booktabs}       
\usepackage{amsfonts}       
\usepackage{nicefrac}       
\usepackage{microtype}      
\usepackage{lipsum}
\usepackage{fancyhdr}       
\usepackage{graphicx}       
\usepackage{xcolor}
\usepackage{cite}

\title{Human vs Objective Evaluation of Colourisation Performance}

\author{
  Seán Mullery \\
  Institute of Technology, Sligo \\
  Sligo\\
  \texttt{mullery.sean@itsligo.ie} \\
   \And
  Paul F. Whelan \\
  Vision Systems Group \\
  Dublin City University \\
  Dublin\\
  \texttt{paul.whelan@dcu.ie} \\
}

\begin{document}




\maketitle

\begin{abstract}
Automatic colourisation of grey-scale images is the process of creating a full-colour image from the grey-scale prior. It is an ill-posed problem, as there are many plausible colourisations for a given grey-scale prior. The current SOTA in auto-colourisation involves image-to-image type Deep Convolutional Neural Networks with Generative Adversarial Networks showing the greatest promise. The end goal of colourisation is to produce full colour images that appear plausible to the human viewer, but human assessment is costly and time consuming. This work assesses how well commonly used objective measures correlate with human opinion. We also attempt to determine what facets of colourisation have the most significant effect on human opinion.
For each of 20 images from the BSD dataset, we create 65 recolourisations made up of local and global changes. Opinion scores are then crowd sourced using the Amazon Mechanical Turk and together with the images this forms an extensible dataset called the Human Evaluated Colourisation Dataset (HECD). While we find statistically significant correlations between human-opinion scores and a small number of objective measures, the strength of the correlations is low. There is also evidence that human observers are most intolerant to an incorrect hue of naturally occurring objects. 
\end{abstract}

\section{Introduction}
The goal of automatic colourisation is to convert grey-scale images to colour images. Colourisation is an ill-posed problem as many plausible colourisations can result from the same grey-scale image. Predicting the exact colour of the original scene is impossible without further prior information from historical sources. A recent trend is to take any natural colour image dataset, convert it to a luminance-chrominance colour space, use the luminance channel as the grey-scale image and use the two chrominance channels as the ground-truth that a deep neural network must predict. This admits only a single ground-truth for each grey-scale image despite other plausible colourisations existing. Objective assessment of a colourisation model's predictions generally relies on distance measures from the ground-truth image.\\
In this paper we wish to answer the following questions.
\begin{itemize}
    \item Do commonly used objective measures of colourisation performance correlate with mean human opinion scores?
    \item Is the ground-truth the perfect colourisation for its grey-scale prior?
    \item Will correction of white-balance of ground-truth images lead to higher opinion score?
    \item Are there any image statistics that all plausible colourisations might have in common?
    
\end{itemize} 

Our key contributions are:
\begin{itemize}
    \item An extensible Human Evaluated Colourisation dataset of recolourisations with matching human-opinion scores to benchmark future objective measures of colourisation. 
    \item An assessment of the correlation between the human-opinion score of colourisation performance and the objective measures used in the colourisation literature.
    \item Analysis and insight into aspects of colourisation that affect the human-opinion score.
    \item An interactive tool to allow other researchers to analyse the HECD dataset and its results - \url{https://github.com/seanmullery/HECD}
\end{itemize} 

\section{How is colourisation performance measured in the literature?}\label{LitReview}

Most colourisation techniques rely on some form of human-visual inspection to determine efficacy, or for comparison to other techniques.  Human-visual inspection can include qualitative analysis \cite{IronyColorByExample2005, Isola2017, WelshTransferingColor:2002:TCG:566654.566576, YatzivFastIMageandVideoColorization, ZhangIE16ColorfulImageColorization, ZhangZIGLYE17UserGuidedColorization, AnImageColorizationWithCNN, SuInstanceAwareColorization, LiAutomaticExample-BasedColorization,  RoyerProbabilisticImageColorizationBMVC2017_85, gorriz2019endtoendColorization, LeeReference-BasedSketchImageColorization, cao2017unsupervisedColorization, YooFewShotColorizationMemoryAugmentedNetworks}, naturalness scoring \cite{AnImageColorizationWithCNN, IizukaLetThereBeColor:2016:LCJ:2897824.2925974, Zhao0SH018Pixel-LevelsemanticsGuidedImageColorization}, user preference between two options \cite{LiAutomaticExample-BasedColorization}, Visual Turing Test (VTT) judged by human \cite{ZhangIE16ColorfulImageColorization, ZhangZIGLYE17UserGuidedColorization, cao2017unsupervisedColorization, PixColorGuadarramaBMVC2017_112}, which of two colourisations best match a reference image's colour \cite{LiAutomaticExample-BasedColorization}, or which, from many images, appears closest to a ground-truth \cite{YooFewShotColorizationMemoryAugmentedNetworks}. \\
Many attempt an objective measure based on absolute pixel value errors, such as RMSE (Root Mean Squared Error) or $L_2$ pixel distance \cite{ZhangIE16ColorfulImageColorization, AnImageColorizationWithCNN, Deshpande7410429LargeScaleAutoColorization, DeshpandeLearningDiverseImageColorization},  MAE (Mean Absolute Error) or $L_1$ pixel distance \cite{gorriz2019endtoendColorization}, and
PSNR (Peak Signal to Noise Ratio) \cite{ZhangZIGLYE17UserGuidedColorization, SuInstanceAwareColorization, gorriz2019endtoendColorization, Zhao0SH018Pixel-LevelsemanticsGuidedImageColorization, Cheng2015DeepC, KimDeepEdgeAwareInteractiveBleeding, OzbulakColorizationCapsule}.  \cite{LeeReference-BasedSketchImageColorization} develop a patch based version of PSNR called SC-PSNR (Semantically Corresponding PSNR), as they wish to compare colour to a semantically similar patch from a reference image. SSIM (Structural Similarity Index Measure) \cite{SSIM}, is used by \cite{SuInstanceAwareColorization, OzbulakColorizationCapsule, ZhaoPixelatedSemanticColorization} and its multi-scale version MS-SSIM \cite{MS-SSIM} is used by \cite{WuRemoteSensingColorizationGAN}.\\
\cite{KimDeepEdgeAwareInteractiveBleeding} developed an objective measure called CDR (Cluster Discrepancy Ratio)  based on SLIC (Simple Linear Iterative Clustering) superpixels \cite{AchantaSLIC}. CDR is formulated by looking at the discrepancy between super-pixel assignment for ground-truth versus colourisation. Similarly,  \cite{Zhao0SH018Pixel-LevelsemanticsGuidedImageColorization} use mean IoU of segmentation results on the PASCAL VOC2012 dataset \cite{pascal-voc-2012}.\\
\cite{WuVividDiverseColorization} use a no-reference measure called colourfulness score \cite{HaslerColorfulness} which incorporates the means and standard deviations of the a* and b* channels of CIEL*a*b* in a parametric model to compute a measure of how colourful the image is. The parameters were learned from data based on psychophysical experiments .\\
\cite{gorriz2019endtoendColorization} and \cite{PixColorGuadarramaBMVC2017_112} 
compare histograms in the a* and b* channels of CIEL*a*b* over a distribution of images.\\
Some methods, \cite{ZhangIE16ColorfulImageColorization, larsson2016learning, Vitoria_2020_WACV},  utilise the concept that colour will assist in classifying objects. Therefore a neural network designed to classify objects using colour images will show a deterioration in performance if inferred with a poorly colourised image. The difference can then be used as a proxy measure for colourisation performance.   \cite{gorriz2019endtoendColorization} compare $L_1$ distance between convolutional features in the VGG19 model \cite{SimonyanZ14aVGG} for ground-truth and colourised samples.
Similarly, \cite{LeeReference-BasedSketchImageColorization} and \cite{WuVividDiverseColorization} use Fr\'echet Inception Distance \cite{HeuselRUNKH17}, which requires comparing the inception score for colourisations versus ground-truth for 50K samples. 
\cite{LpipsZhang}, developed a perception measure based on the features of deep neural networks called the Learned Perceptual Image Patch Similarity (LPIPS) metric, and this has also been used for the measure of colourisation in \cite{SuInstanceAwareColorization, YooFewShotColorizationMemoryAugmentedNetworks, KimDeepEdgeAwareInteractiveBleeding}.\\
The work of \cite{Anwar2020ImageCA} is the only work we have found which attempt a dataset that is specifically designed for colourisation. Their dataset is designed with the idea of restricting synthetic objects or natural objects such as flowers that may have a wide distribution of plausible colours. Instead they include only natural objects that would be considered to have a narrow distribution of plausible colours such as specific types of fruit and vegetables. The images contain only a single object type, against a white background. There are 20 categories and 723 images in all. They then use PSNR, SSIM, PCQI (Patch-based contrast quality index), and UIQM (Underwater Image Quality Metric) to test out SOTA algorithms on their dataset. As is explained in section \ref{datasetSec}, our dataset design is considerably different and includes human evaluation data for each image.

\section{The Human Evaluated Colourisation Dataset} \label{datasetSec}
The Human Evaluated Colourisation Dataset is based on 20 images from the Berkeley Segmentation Dataset \cite{BSDMartinFTM01}. From each of these 20, 65 images are created that differ in colour from the original. While we attempt to make changes that will be interpretable later, our primary objective is to have many different colour versions for human evaluation to allow appropriate comparison to objective measures. In total, $65\times20=1300$ and 20 original images will result in a total of 1320 images in the set. The BSD set was chosen, as it has a variety of natural images and multiple human segmentations of each. The segmentations, in many cases, segment colour sections, allowing the alteration of the colour of specific sections without modification of the rest of the image. The original image will be referred to as the ground-truth from here on. The following changes are made to the ground-truth to create the HECD dataset.\\
The first recolour modification is to auto-white-balance correct the 20 ground-truth images in Photoshop \cite{Photoshop}, creating 20 new images. To ensure that Photoshop has not changed the L*-channel, the L*-channel is replaced with the ground-truth to ensure that only changes are made to a*b*. While the a*b* channels are close to perceptually uniform, they are not very intuitive, so a reformulation of these channels to hue and chroma channels is used via the equations of \cite{fairchild.ch10}.
\begin{equation}
c=\sqrt{\left(a^{* 2}+b^{*2} \right)}
\label{cab}
\end{equation}
\begin{equation}
h=\tan ^{-1}\left(b^{*} / a^{*}\right)
\label{hab}
\end{equation}
Where $h$ is hue, and $c$ is chroma. We now proceed to make the following global changes to the 40 images (20 ground-truth + 20 WB corrected). The changes below are arbitrary as there is no prior work to guide sample spacing or types of parameters:
\begin{itemize}
    \item Alter intensity value of chroma by $\pm2\sigma, \pm1\sigma$ of the chroma of the image ($4\times 40 = 160$ images).
    \item Alter contrast of chroma by $\frac{1}{4}, \frac{1}{2}, 2, 4$ ($4\times 40 = 160$ images).
    \item Shift (offset registration) the a*b* channels spatially relative to the L*-channel by $0.01, 0.02, 0.03, 0.04$ of the width and height of the image ($4\times 40 = 160$ images). The edges that had no donor pixels, just retain their original value. 
    \item Collect some SOTA colourisation algorithms' predictions of colour given the L*-channels of the 20 ground-truth images. The choice of which SOTA methods to include was based on availability of implementation and ability to accept the BSD image sizes without modification. The six methods chosen were, Photoshop Neural Filter \cite{Photoshop}, Deoldify \cite{MyHeritage}, available on MyHeritage.com, \cite{larsson2016learning},  \cite{ZhangIE16ColorfulImageColorization},  \cite{ZhangZIGLYE17UserGuidedColorization} (using straight-forward inference with no user guidance), and  \cite{IizukaLetThereBeColor:2016:LCJ:2897824.2925974}. We also replace these L*-channels with the ground-truth L*-channel in case any of the SOTA algorithms alter the L*-channel as part of their processing pipeline. $6\times20 = 120$ images.
\end{itemize}

In addition to the global changes, we also introduce some local changes. For each of the 40 images, we choose either a single segment or multiple segments that are of the same colour and make the following modifications to just the chosen segment(s).

\begin{itemize}
    \item For the segment we alter the intensity of the chroma by $\pm2\sigma, \pm1\sigma$ of the chroma of the image ($4\times 40 = 160$ images).
    \item Hue is not a magnitude space; you cannot have an absence of hue or more/less hue, and all hues are equally important. Therefore, statistics like mean and standard deviation are not meaningful in the hue channel. We alter the hue of the segment in a logarithmic fashion so that we can get better resolution in results closely surrounding the reference hue but still cover the full space of hue values without the cost of sampling all 256 hue values. Future extensions could more tightly sample the whole space.  With the hue from Equation \ref{hab}  forming a circular space  $\in [0,255]$ we make the following alterations from the reference hue. $\pm 2,\pm4,\pm8,\pm16,\pm32,\pm64,$ and  $128$ ($\pm128$ results in the same change). ($13\times 40 = 520$ images). 
\end{itemize}

While this is a small dataset by current standards, it has been designed with extensibility in mind. The arbitrary modifications above were chosen to return the most information for the available resources. More ground-truth images and more recolour modifications along with tighter sampling between modification types, could be added in the future by collecting data in a manner consistent with that given in section \ref{collectingSec}.

\begin{figure}[t]

	\includegraphics[width=\textwidth]{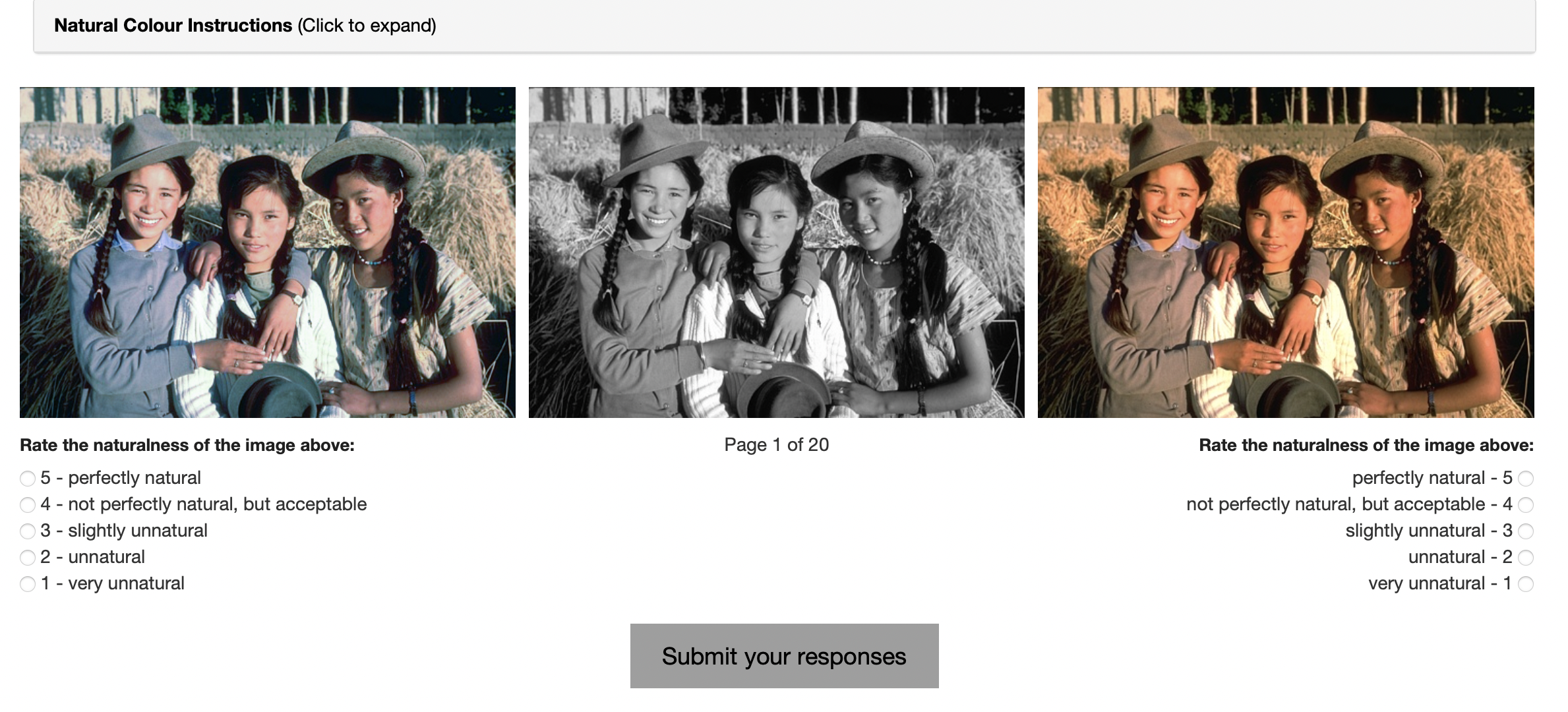}
	\caption{Each survey question displays three images. In the middle is the L*-channel, which is common to the three images. On either side are the ground-truth and a recolourisation. The participants are not informed that one image is the ground-truth and it could appear on either right or left with equal probability. The participant must respond to both before continuing.}
	
	\label{LabelfigSur}
 \end{figure}

\section{Collecting the data.}\label{collectingSec}
The Amazon Mechanical Turk was used to assess human opinion on colourisation. Ethics approval was obtained in accordance with the University's Research Ethics Committee guidelines. Each assessment consisted of three images appearing on the screen simultaneously: the L*-channel (in the middle) and a colourisation on each side. One of the colourisations is the ground-truth colour image, and the other is one of the modifications described in section \ref{datasetSec}, see Figure   \ref{LabelfigSur}. In this manner, all scores have a control in common. The observer is not informed that one image is the ground-truth and the positions vary in a pseudo-random manner so that there is an equal likelihood that the ground-truth could be on the left or the right. The user is asked to score the two colour images on naturalness (how much the colour looks like it would appear in real life). The scores are from 1-5 on an ordinal scale. 
Each observer rates 20 pairs. They see each of the 20 ground-truth images in the dataset and a recolourised version. For any set of 20, the type of recolourisation is pseudo-random, so the user does not become accustomed to the type of colour change. As there are 65 recolour versions, 65 surveys of 20 comparisons are created for 1300 responses in total. While each survey is pseudo-random internally, the actual survey is identical for each observer that responds to it. We do not allow a participant to respond to a unique survey twice. In general, we allow participants to complete only one survey. A small number completed two different surveys (19 participants). This is not a problem, but if participants were allowed to do many surveys in a short period, it could lead to non-naivete, with the participant learning that specific colour versions appear in all surveys, i.e. they may learn to recognise the ground-truth and be biased towards awarding it the higher of the two scores. In all, there were 1281 participants. Twenty participants completed each survey. Twenty-nine incomplete surveys were not used but also not counted in the total 1300 complete surveys. In surveys with more than one response for a pair of images (respondent used the back button in the browser), the final result was used on the assumption that this is what the respondent intended. There were 25 surveys where the user gave the same value for all answers (straight-lining) and 15 where the respondent gave the same number for the two images under consideration in all 20 comparisons; these were removed from the data, leaving 1260 complete surveys.

\section{Processing the raw numbers} \label{ProcRawSec}
As the ground-truth image was used as the control, it is the difference between the score for the ground-truth and the recoloured image in which we are interested. However, account for differences in individual participants that may bias the results still needs to be taken. One participant may score all pairs lower than another participant, with all else equal. As the ordinal values for scoring and the gaps between them are subjective, two participants who perceive the same difference between two images may still give a larger/smaller difference in scores compared to each other. Differences in viewing equipment/environment may also have systematic effects between two respondents. For this reason, it is necessary to consider the trend for the participant as a whole over the 20 image pairs to which they respond. The method of  \cite{Sheikh1709988} can then be used to calculate the difference for each pair.
\begin{equation}
d_{ij}=r_{ij} - r_{iref(j)}
\end{equation}
where $r_{ij}$ is the raw score for the $i$-th participant and $j$-th image, and $r_{iref(j)}$ denotes the raw quality
score assigned by the $i$-th participant to the reference image corresponding to the $j$-th recolourised image. The
raw difference scores $d_{ij}$ for the $i$-th participant and $j$-th image are converted into Z-scores.

\begin{equation}
z_{ij}=(d_{ij}-\bar{d_i})/\sigma_i
\end{equation}

Where $\bar{d_i}$ is the mean of the raw difference scores over all of the images ranked by participant $i$, and $\sigma_i$ is the standard deviation of the differences for participant $i$. 
$z_{ij}$ then represents a score for an image $j$ by participant $i$. In most cases, in this document, the score for an image $j$ is given as the mean over all the participants that responded to it. Because the ground-truth is used as the control and the processing is based on the statistics of the participants, the ground-truth images are all considered of equal quality. When their z-scores are calculated and averaged, they all come to the same value. As there were more recolourisations that scored lower than the ground-truth than those scoring higher, the average score for the ground-truth will have a positive non-zero value.

\section{Results}

\subsection{How do objective measures correlate with human opinion?}
As outlined in section \ref{LitReview}, colourisation researchers have attempted to use many different types of objective measures to assess the quality of colourisations. We test if the human scores correlate with the commonly used objective measures. As ordinal data is used,  two rank-order correlation measures will be used to examine the rank order correlation of the results, namely Spearman-r \cite{Spearman}, see table \ref{spearmanAll} and Kendall-tau \cite{kendall1938measure}, see table \ref{kendallAll}. The shaded values, in the tables, represent values where the p-value of the rank-order correlation was less than 0.05, indicating statistical significance. We test against SSIM \cite{SSIM}, MS-SSIM \cite{MS-SSIM}, MSE, RMSE/$L_2$, MAE/$L_1$, Colourfulness and Colourfulness Difference \cite{HaslerColorfulness}, PSNR, CDR \cite{KimDeepEdgeAwareInteractiveBleeding} and LPIPS \cite{LpipsZhang} for both VGG \cite{SimonyanZ14aVGG} and Alexnet \cite{Alexnet}. We cannot test FID \cite{HeuselRUNKH17}, or SC-PSNR \cite{LeeReference-BasedSketchImageColorization}, as they require different data than we have used for the surveys. Where possible, we use established libraries for the measures. SKImage \cite{van2014scikit} for SSIM and PSNR, Sewar \cite{sewar} for MS-SSIM, RMSE ($L_2$), and MSE, and SKLearn \cite{pedregosa2011scikit} for MAE ($L_1$). Colourfulness and Colourfulness-Difference are developed from the details in \cite{HaslerColorfulness}. CDR is developed from the details in \cite{KimDeepEdgeAwareInteractiveBleeding} and relies on SKImage's SLIC library. We also test in three different colour spaces where the method is not specific to a particular colour space, namely a*b*, hc (see Equations \ref{cab} and \ref{hab}) and RGB. a*b* and hc do not include the L*-channel in the comparison as L* is common in all pairings. RGB incorporates the L*-channel but in a different formulation. We can see from tables \ref{spearmanAll} and \ref{kendallAll} that MS-SSIM, when used with either a*b* or RGB, has the strongest correlation with human judgement. Standard SSIM with a*b*  is the only other that has a statistically significant correlation for all images. hc seems to be a poor space in which to use any of the objective measures despite most of the changes in our dataset being made in this formulation, see section \ref{datasetSec}.
Even for the top performer, MS-SSIM with a*b*, the correlation for the complete set with Spearman is 0.567 and Kendall is 0.389. In general, values above 0.7 are considered ``Strong Correlation", so none of the objective measures meets that threshold, though a small number do reach this for an individual image. In short, the objective measures employed in the literature do not work well for colourisation. There is scope here for a more targeted objective measure, and our HECD dataset is publicly available to help in this search.

\begin{table*}
\centering
\caption{Spearman rank order correlation for all reference images individually and all combined. The numbers represent the Spearman-r value, and the shaded numbers are those that are statistically significant with a p-value $<0.05$. The best performer in each row is shown in bold. }
\label{spearmanAll}
\resizebox{1.0\textwidth}{!}{
\begin{tabular}{llllllllllllllllllll}
\hline
\rotatebox{90}{GT FileName} & \rotatebox{90}{SSIM (a*b*)}$ \big\uparrow$& \rotatebox{90}{SSIM (hc)} $ \big\uparrow$ & \rotatebox{90}{SSIM (rgb)} $ \big\uparrow$ & \rotatebox{90}{MS-SSIM (a*b*)}$ \big\uparrow$ & \rotatebox{90}{MS-SSIM (hc)}$ \big\uparrow$ & \rotatebox{90}{MS-SSIM (rgb)} $ \big\uparrow$ & \rotatebox{90}{MSE (a*b*)} $\big\downarrow$& \rotatebox{90}{MSE-tau (hc)} $\big\downarrow$& \rotatebox{90}{RMSE (a*b*)} $\big\downarrow$& \rotatebox{90}{RMSE (hc)} $\big\downarrow$& \rotatebox{90}{MAE (a*b*)} $\big\downarrow$& \rotatebox{90}{MAE (hc)} $\big\downarrow$ & \rotatebox{90}{Colourfulness} $ \big\uparrow$& \rotatebox{90}{Colourfulness-dif} $ \big\downarrow$ &     \rotatebox{90}{psnr-ab} $ \big\uparrow$&     \rotatebox{90}{psnr-hc} $ \big\uparrow$& \rotatebox{90}{CDR} $\big\downarrow$& \rotatebox{90}{lpips-vgg} $ \big\downarrow$&   \rotatebox{90}{lpips-alex} $ \big\downarrow$ \\
\hline
             015004\_gt.jpg &          \colorbox{pink}{0.298} &                       0.165 &                        0.165 &  \textbf{\colorbox{pink}{0.444}} &          \colorbox{pink}{0.242} &          \colorbox{pink}{0.399} &      \colorbox{pink}{-0.294} &                      0.118 &       \colorbox{pink}{-0.292} &                       0.123 &                        0.198 &                     -0.029 &                          -0.228 &                              -0.228 &          \colorbox{pink}{0.367} &                           0.115 & \colorbox{pink}{-0.261} &    \colorbox{pink}{-0.291} &          \colorbox{pink}{-0.352} \\
             022090\_gt.jpg &          \colorbox{pink}{0.303} &                       0.022 &                        0.022 &            \colorbox{pink}{0.29} &                           0.091 & \textbf{\colorbox{pink}{0.331}} &                       -0.121 &                       0.06 &                        -0.136 &                       0.062 &                       -0.183 &                     -0.082 &                          -0.078 &                              -0.078 &                           0.096 &                          -0.078 &                  -0.178 &                     -0.114 &                           -0.125 \\
             022093\_gt.jpg &          \colorbox{pink}{0.364} &                       0.189 &                        0.189 &  \textbf{\colorbox{pink}{0.468}} &                           0.179 &          \colorbox{pink}{0.377} &                       -0.225 &    \colorbox{pink}{-0.296} &                        -0.211 &     \colorbox{pink}{-0.271} &      \colorbox{pink}{-0.437} &                     -0.124 &          \colorbox{pink}{0.411} &              \colorbox{pink}{0.411} &          \colorbox{pink}{0.374} &                          -0.075 & \colorbox{pink}{-0.295} &                     -0.165 &          \colorbox{pink}{-0.267} \\
             024004\_gt.jpg &          \colorbox{pink}{0.498} &      \colorbox{pink}{0.504} &       \colorbox{pink}{0.504} &  \textbf{\colorbox{pink}{0.598}} &          \colorbox{pink}{0.536} &          \colorbox{pink}{0.541} &      \colorbox{pink}{-0.401} &                      -0.23 &       \colorbox{pink}{-0.434} &                      -0.022 &      \colorbox{pink}{-0.459} &                     -0.145 &                           0.098 &                               0.098 &          \colorbox{pink}{0.512} &          \colorbox{pink}{0.259} & \colorbox{pink}{-0.457} &    \colorbox{pink}{-0.538} &          \colorbox{pink}{-0.502} \\
             025098\_gt.jpg &           \colorbox{pink}{0.37} &                       0.193 &                        0.193 &           \colorbox{pink}{0.457} &                           0.099 &          \colorbox{pink}{0.487} &      \colorbox{pink}{-0.363} &                     -0.088 &        \colorbox{pink}{-0.41} &                      -0.032 &                        0.055 &     \colorbox{pink}{0.457} &                           0.018 &                               0.018 & \textbf{\colorbox{pink}{0.568}} &                           0.069 &  \colorbox{pink}{-0.34} &    \colorbox{pink}{-0.369} &          \colorbox{pink}{-0.395} \\
             046076\_gt.jpg &          \colorbox{pink}{0.418} &                       0.062 &                        0.062 &  \textbf{\colorbox{pink}{0.547}} &                           0.042 &          \colorbox{pink}{0.441} &                       -0.194 &    \colorbox{pink}{-0.331} &                        -0.209 &     \colorbox{pink}{-0.262} &                        0.126 &                       0.16 &                           0.125 &                               0.125 &          \colorbox{pink}{0.402} &                          -0.101 &  \colorbox{pink}{-0.27} &    \colorbox{pink}{-0.314} &          \colorbox{pink}{-0.454} \\
             056028\_gt.jpg &          \colorbox{pink}{0.283} &                       0.193 &                        0.193 &            \colorbox{pink}{0.34} &                           0.139 &          \colorbox{pink}{0.302} &                       -0.219 &                     -0.021 &       \colorbox{pink}{-0.255} &                       0.093 &                        -0.21 &                      0.179 &                          -0.195 &                              -0.195 &                            0.23 &                           0.035 & \colorbox{pink}{-0.243} &    \colorbox{pink}{-0.384} & \textbf{\colorbox{pink}{-0.368}} \\
             065019\_gt.jpg &          \colorbox{pink}{0.443} &        \colorbox{pink}{0.6} &         \colorbox{pink}{0.6} &           \colorbox{pink}{0.386} & \textbf{\colorbox{pink}{0.636}} &          \colorbox{pink}{0.444} &      \colorbox{pink}{-0.426} &    \colorbox{pink}{-0.282} &        \colorbox{pink}{-0.43} &     \colorbox{pink}{-0.285} &      \colorbox{pink}{-0.308} &    \colorbox{pink}{-0.428} &                           0.215 &                               0.215 &          \colorbox{pink}{0.346} &          \colorbox{pink}{0.538} & \colorbox{pink}{-0.533} &    \colorbox{pink}{-0.399} &          \colorbox{pink}{-0.337} \\
             078019\_gt.jpg &          \colorbox{pink}{0.577} &      \colorbox{pink}{0.628} &       \colorbox{pink}{0.628} &            \colorbox{pink}{0.64} &           \colorbox{pink}{0.66} &          \colorbox{pink}{0.585} &      \colorbox{pink}{-0.617} &                      0.212 &        \colorbox{pink}{-0.61} &       \colorbox{pink}{0.34} &                       -0.112 &     \colorbox{pink}{0.504} &                          -0.238 &                              -0.238 &          \colorbox{pink}{0.561} & \textbf{\colorbox{pink}{0.685}} & \colorbox{pink}{-0.539} &    \colorbox{pink}{-0.602} &          \colorbox{pink}{-0.589} \\
             118020\_gt.jpg &          \colorbox{pink}{0.511} &       \colorbox{pink}{0.33} &        \colorbox{pink}{0.33} &  \textbf{\colorbox{pink}{0.594}} &          \colorbox{pink}{0.281} &          \colorbox{pink}{0.582} &      \colorbox{pink}{-0.517} &                     -0.018 &       \colorbox{pink}{-0.508} &                       0.023 &                       -0.075 &                      0.143 &         \colorbox{pink}{-0.321} &             \colorbox{pink}{-0.321} &          \colorbox{pink}{0.526} &                           0.043 & \colorbox{pink}{-0.448} &    \colorbox{pink}{-0.539} &          \colorbox{pink}{-0.513} \\
             118035\_gt.jpg &          \colorbox{pink}{0.673} &      \colorbox{pink}{0.489} &       \colorbox{pink}{0.489} &  \textbf{\colorbox{pink}{0.694}} &          \colorbox{pink}{0.626} &          \colorbox{pink}{0.617} &      \colorbox{pink}{-0.612} &    \colorbox{pink}{-0.315} &       \colorbox{pink}{-0.607} &     \colorbox{pink}{-0.254} &                        0.177 &     \colorbox{pink}{0.445} &         \colorbox{pink}{-0.472} &             \colorbox{pink}{-0.472} &          \colorbox{pink}{0.565} &          \colorbox{pink}{0.651} &  \colorbox{pink}{-0.51} &    \colorbox{pink}{-0.596} &          \colorbox{pink}{-0.569} \\
             140075\_gt.jpg &          \colorbox{pink}{0.627} &      \colorbox{pink}{0.436} &       \colorbox{pink}{0.436} &  \textbf{\colorbox{pink}{0.651}} &          \colorbox{pink}{0.386} &          \colorbox{pink}{0.627} &        \colorbox{pink}{-0.6} &    \colorbox{pink}{-0.325} &       \colorbox{pink}{-0.611} &     \colorbox{pink}{-0.252} &                       -0.176 &                     -0.089 &                           0.157 &                               0.157 &          \colorbox{pink}{0.405} &           \colorbox{pink}{0.29} & \colorbox{pink}{-0.438} &    \colorbox{pink}{-0.586} &          \colorbox{pink}{-0.545} \\
             151087\_gt.jpg &          \colorbox{pink}{0.633} &       \colorbox{pink}{0.57} &        \colorbox{pink}{0.57} &   \textbf{\colorbox{pink}{0.65}} &          \colorbox{pink}{0.597} &          \colorbox{pink}{0.618} &      \colorbox{pink}{-0.608} &                      0.183 &       \colorbox{pink}{-0.608} &                        0.19 &                          0.0 &                      0.061 &                          -0.023 &                              -0.023 &          \colorbox{pink}{0.611} &          \colorbox{pink}{0.299} & \colorbox{pink}{-0.579} &    \colorbox{pink}{-0.583} &          \colorbox{pink}{-0.559} \\
             153093\_gt.jpg &          \colorbox{pink}{0.561} &      \colorbox{pink}{0.458} &       \colorbox{pink}{0.458} &           \colorbox{pink}{0.647} &                           0.206 &          \colorbox{pink}{0.621} &       \colorbox{pink}{-0.28} &                      0.029 &       \colorbox{pink}{-0.354} &                       0.201 &                        0.043 &     \colorbox{pink}{0.275} &                          -0.112 &                              -0.112 &          \colorbox{pink}{0.394} &                           0.106 & \colorbox{pink}{-0.728} &    \colorbox{pink}{-0.717} & \textbf{\colorbox{pink}{-0.709}} \\
             187029\_gt.jpg & \textbf{\colorbox{pink}{0.545}} &      \colorbox{pink}{0.338} &       \colorbox{pink}{0.338} &           \colorbox{pink}{0.522} &           \colorbox{pink}{0.39} &          \colorbox{pink}{0.541} &      \colorbox{pink}{-0.314} &                      0.153 &       \colorbox{pink}{-0.354} &                       0.172 &                        0.025 &      \colorbox{pink}{0.48} &          \colorbox{pink}{0.273} &              \colorbox{pink}{0.273} &          \colorbox{pink}{0.538} &                           0.053 & \colorbox{pink}{-0.513} &    \colorbox{pink}{-0.527} &          \colorbox{pink}{-0.529} \\
             198023\_gt.jpg &          \colorbox{pink}{0.539} &      \colorbox{pink}{0.327} &       \colorbox{pink}{0.327} &           \colorbox{pink}{0.648} &                          -0.034 &          \colorbox{pink}{0.667} &       \colorbox{pink}{-0.58} &     \colorbox{pink}{0.429} &       \colorbox{pink}{-0.589} &      \colorbox{pink}{0.433} &                        0.218 &     \colorbox{pink}{0.468} &         \colorbox{pink}{-0.256} &             \colorbox{pink}{-0.256} &          \colorbox{pink}{0.608} &                          -0.169 & \colorbox{pink}{-0.694} &    \colorbox{pink}{-0.658} & \textbf{\colorbox{pink}{-0.726}} \\
             239096\_gt.jpg &          \colorbox{pink}{0.578} &      \colorbox{pink}{0.585} &       \colorbox{pink}{0.585} &           \colorbox{pink}{0.532} &          \colorbox{pink}{0.438} &          \colorbox{pink}{0.583} &      \colorbox{pink}{-0.501} &    \colorbox{pink}{-0.258} &       \colorbox{pink}{-0.509} &                      -0.138 &                       -0.231 &                     -0.002 &                           0.012 &                               0.012 &          \colorbox{pink}{0.455} &                           0.192 & \colorbox{pink}{-0.489} &    \colorbox{pink}{-0.604} & \textbf{\colorbox{pink}{-0.566}} \\
             242078\_gt.jpg &          \colorbox{pink}{0.509} &      \colorbox{pink}{0.424} &       \colorbox{pink}{0.424} &           \colorbox{pink}{0.531} &          \colorbox{pink}{0.351} &          \colorbox{pink}{0.584} &      \colorbox{pink}{-0.441} &    \colorbox{pink}{-0.299} &       \colorbox{pink}{-0.422} &                      -0.192 &                       -0.072 &                      0.032 &                          -0.057 &                              -0.057 &                           0.135 &                           0.161 & \colorbox{pink}{-0.524} &    \colorbox{pink}{-0.736} & \textbf{\colorbox{pink}{-0.725}} \\
             323016\_gt.jpg &          \colorbox{pink}{0.642} &      \colorbox{pink}{0.247} &       \colorbox{pink}{0.247} &  \textbf{\colorbox{pink}{0.724}} &          \colorbox{pink}{0.332} &            \colorbox{pink}{0.7} &      \colorbox{pink}{-0.531} &                      0.172 &       \colorbox{pink}{-0.515} &                       0.118 &                        0.082 &    \colorbox{pink}{-0.315} &         \colorbox{pink}{-0.314} &             \colorbox{pink}{-0.314} &          \colorbox{pink}{0.496} &          \colorbox{pink}{0.315} & \colorbox{pink}{-0.481} &     \colorbox{pink}{-0.57} &          \colorbox{pink}{-0.653} \\
             376001\_gt.jpg &          \colorbox{pink}{0.318} &      \colorbox{pink}{0.279} &       \colorbox{pink}{0.279} &           \colorbox{pink}{0.315} &                           0.234 & \textbf{\colorbox{pink}{0.448}} &                       -0.238 &                     -0.063 &       \colorbox{pink}{-0.245} &                      -0.068 &                        0.085 &                     -0.057 &                          -0.039 &                              -0.039 &          \colorbox{pink}{0.378} &                           0.129 & \colorbox{pink}{-0.399} &    \colorbox{pink}{-0.394} &          \colorbox{pink}{-0.424} \\
             \hline
                        All &          \colorbox{pink}{0.492} &      \colorbox{pink}{0.271} &       \colorbox{pink}{0.422} &  \textbf{\colorbox{pink}{0.567}} &           \colorbox{pink}{0.29} &          \colorbox{pink}{0.549} &      \colorbox{pink}{-0.416} &                      0.015 &       \colorbox{pink}{-0.434} &      \colorbox{pink}{0.091} &      \colorbox{pink}{-0.135} &                      0.034 &                          -0.022 &              \colorbox{pink}{0.071} &           \colorbox{pink}{0.45} &          \colorbox{pink}{0.068} & \colorbox{pink}{-0.461} &    \colorbox{pink}{-0.474} &          \colorbox{pink}{-0.495} \\
\hline
\end{tabular}}
\end{table*}


\begin{table*}[!t]

\centering
\caption{Kendall rank order correlation for all reference images individually and all combined. The numbers represent the Kendall tau value, and the shaded numbers are those that are statistically significant with a p-value $<0.05$. The best performer in each row is shown in bold. }
\label{kendallAll}
\resizebox{1.0\textwidth}{!}{
\begin{tabular}{llllllllllllllllllll}
\hline
\rotatebox{90}{GT FileName} & \rotatebox{90}{SSIM (a*b*)}$ \big\uparrow$& \rotatebox{90}{SSIM (hc)} $ \big\uparrow$ & \rotatebox{90}{SSIM (rgb)} $ \big\uparrow$ & \rotatebox{90}{MS-SSIM (a*b*)}$ \big\uparrow$ & \rotatebox{90}{MS-SSIM (hc)}$ \big\uparrow$ & \rotatebox{90}{MS-SSIM (rgb)} $ \big\uparrow$ & \rotatebox{90}{MSE (a*b*)} $\big\downarrow$& \rotatebox{90}{MSE (hc)} $\big\downarrow$& \rotatebox{90}{RMSE (a*b*)} $\big\downarrow$& \rotatebox{90}{RMSE (hc)} $\big\downarrow$& \rotatebox{90}{MAE (a*b*)} $\big\downarrow$& \rotatebox{90}{MAE (hc)} $\big\downarrow$ & \rotatebox{90}{Colourfulness} $ \big\uparrow$& \rotatebox{90}{Colourfulness-dif} $ \big\downarrow$ &     \rotatebox{90}{psnr-ab} $ \big\uparrow$&     \rotatebox{90}{psnr-hc} $ \big\uparrow$& \rotatebox{90}{CDR} $\big\downarrow$& \rotatebox{90}{lpips-vgg} $ \big\downarrow$&   \rotatebox{90}{lpips-alex} $ \big\downarrow$\\
\hline
             015004\_gt.jpg &          \colorbox{pink}{0.205} &                         0.098 &                          0.098 &    \textbf{\colorbox{pink}{0.299}} &                            0.158 &             \colorbox{pink}{0.28} &        \colorbox{pink}{-0.194} &                        0.075 &         \colorbox{pink}{-0.191} &                         0.083 &                          0.126 &                       -0.014 &                            -0.161 &                                 0.161 &          \colorbox{pink}{0.247} &                           0.076 &  \colorbox{pink}{-0.19} &       \colorbox{pink}{-0.201} &          \colorbox{pink}{-0.234} \\
             022090\_gt.jpg &          \colorbox{pink}{0.205} &                         0.016 &                          0.016 &             \colorbox{pink}{0.197} &                            0.071 &   \textbf{\colorbox{pink}{0.215}} &                         -0.093 &                        0.051 &                          -0.104 &                         0.055 &                         -0.123 &                       -0.072 &                            -0.055 &                                 0.055 &                           0.075 &                          -0.058 &                  -0.122 &                        -0.083 &                           -0.085 \\
             022093\_gt.jpg &          \colorbox{pink}{0.245} &                         0.128 &                          0.128 &     \textbf{\colorbox{pink}{0.31}} &                            0.125 &            \colorbox{pink}{0.248} &                         -0.149 &      \colorbox{pink}{-0.201} &                          -0.145 &       \colorbox{pink}{-0.187} &        \colorbox{pink}{-0.306} &                       -0.087 &            \colorbox{pink}{0.294} &               \colorbox{pink}{-0.294} &          \colorbox{pink}{0.241} &                          -0.032 & \colorbox{pink}{-0.217} &                        -0.114 &          \colorbox{pink}{-0.192} \\
             024004\_gt.jpg &           \colorbox{pink}{0.35} &        \colorbox{pink}{0.351} &         \colorbox{pink}{0.351} &    \textbf{\colorbox{pink}{0.425}} &           \colorbox{pink}{0.371} &             \colorbox{pink}{0.38} &        \colorbox{pink}{-0.273} &      \colorbox{pink}{-0.169} &           \colorbox{pink}{-0.3} &                        -0.025 &        \colorbox{pink}{-0.314} &                       -0.085 &                              0.07 &                                 -0.07 &          \colorbox{pink}{0.344} &          \colorbox{pink}{0.186} & \colorbox{pink}{-0.316} &       \colorbox{pink}{-0.376} &          \colorbox{pink}{-0.346} \\
             025098\_gt.jpg &           \colorbox{pink}{0.27} &                         0.138 &                          0.138 &             \colorbox{pink}{0.333} &                            0.072 &            \colorbox{pink}{0.342} &        \colorbox{pink}{-0.251} &                       -0.065 &         \colorbox{pink}{-0.283} &                        -0.032 &                          0.031 &       \colorbox{pink}{0.316} &                             0.012 &                                -0.012 & \textbf{\colorbox{pink}{0.398}} &                           0.049 & \colorbox{pink}{-0.243} &       \colorbox{pink}{-0.254} &          \colorbox{pink}{-0.271} \\
             046076\_gt.jpg &          \colorbox{pink}{0.276} &                         0.063 &                          0.063 &    \textbf{\colorbox{pink}{0.378}} &                            0.044 &            \colorbox{pink}{0.295} &                         -0.128 &      \colorbox{pink}{-0.235} &                          -0.141 &       \colorbox{pink}{-0.191} &                          0.096 &                        0.123 &                             0.101 &                                -0.101 &          \colorbox{pink}{0.273} &                           -0.02 & \colorbox{pink}{-0.179} &        \colorbox{pink}{-0.21} &          \colorbox{pink}{-0.303} \\
             056028\_gt.jpg &           \colorbox{pink}{0.18} &                         0.133 &                          0.133 &             \colorbox{pink}{0.216} &                            0.101 &            \colorbox{pink}{0.198} &                         -0.152 &                       -0.013 &         \colorbox{pink}{-0.174} &                         0.065 &                         -0.153 &                        0.131 &                            -0.138 &                                 0.138 &                           0.142 &                           0.027 &                  -0.159 &       \colorbox{pink}{-0.265} & \textbf{\colorbox{pink}{-0.253}} \\
             065019\_gt.jpg &          \colorbox{pink}{0.319} &        \colorbox{pink}{0.423} &         \colorbox{pink}{0.423} &             \colorbox{pink}{0.276} &  \textbf{\colorbox{pink}{0.458}} &            \colorbox{pink}{0.327} &         \colorbox{pink}{-0.29} &      \colorbox{pink}{-0.198} &         \colorbox{pink}{-0.293} &       \colorbox{pink}{-0.189} &        \colorbox{pink}{-0.193} &      \colorbox{pink}{-0.269} &                             0.152 &                                -0.152 &          \colorbox{pink}{0.241} &          \colorbox{pink}{0.373} & \colorbox{pink}{-0.389} &       \colorbox{pink}{-0.302} &          \colorbox{pink}{-0.262} \\
             078019\_gt.jpg &          \colorbox{pink}{0.409} &        \colorbox{pink}{0.456} &         \colorbox{pink}{0.456} &             \colorbox{pink}{0.461} &           \colorbox{pink}{0.483} &            \colorbox{pink}{0.414} &         \colorbox{pink}{-0.43} &                        0.141 &         \colorbox{pink}{-0.427} &        \colorbox{pink}{0.227} &                         -0.097 &       \colorbox{pink}{0.357} &           \colorbox{pink}{-0.179} &                \colorbox{pink}{0.179} &          \colorbox{pink}{0.386} & \textbf{\colorbox{pink}{0.509}} &  \colorbox{pink}{-0.38} &       \colorbox{pink}{-0.437} &          \colorbox{pink}{-0.432} \\
             118020\_gt.jpg &           \colorbox{pink}{0.35} &        \colorbox{pink}{0.229} &         \colorbox{pink}{0.229} &    \textbf{\colorbox{pink}{0.412}} &           \colorbox{pink}{0.215} &            \colorbox{pink}{0.401} &        \colorbox{pink}{-0.345} &                       -0.028 &         \colorbox{pink}{-0.345} &                           0.0 &                         -0.055 &                          0.1 &            \colorbox{pink}{-0.23} &                 \colorbox{pink}{0.23} &          \colorbox{pink}{0.356} &                           0.055 & \colorbox{pink}{-0.309} &       \colorbox{pink}{-0.387} &          \colorbox{pink}{-0.359} \\
             118035\_gt.jpg &          \colorbox{pink}{0.476} &        \colorbox{pink}{0.325} &         \colorbox{pink}{0.325} &    \textbf{\colorbox{pink}{0.485}} &           \colorbox{pink}{0.433} &            \colorbox{pink}{0.447} &        \colorbox{pink}{-0.416} &      \colorbox{pink}{-0.227} &          \colorbox{pink}{-0.41} &       \colorbox{pink}{-0.175} &                          0.119 &       \colorbox{pink}{0.303} &           \colorbox{pink}{-0.334} &                \colorbox{pink}{0.334} &          \colorbox{pink}{0.381} &          \colorbox{pink}{0.448} & \colorbox{pink}{-0.362} &       \colorbox{pink}{-0.401} &          \colorbox{pink}{-0.404} \\
             140075\_gt.jpg &          \colorbox{pink}{0.448} &        \colorbox{pink}{0.305} &         \colorbox{pink}{0.305} &    \textbf{\colorbox{pink}{0.473}} &           \colorbox{pink}{0.267} &            \colorbox{pink}{0.441} &        \colorbox{pink}{-0.411} &      \colorbox{pink}{-0.231} &         \colorbox{pink}{-0.424} &       \colorbox{pink}{-0.183} &                         -0.124 &                       -0.066 &                             0.098 &                                -0.098 &          \colorbox{pink}{0.284} &          \colorbox{pink}{0.196} & \colorbox{pink}{-0.302} &       \colorbox{pink}{-0.414} &           \colorbox{pink}{-0.37} \\
             151087\_gt.jpg &          \colorbox{pink}{0.437} &         \colorbox{pink}{0.37} &          \colorbox{pink}{0.37} &    \textbf{\colorbox{pink}{0.451}} &           \colorbox{pink}{0.404} &             \colorbox{pink}{0.42} &        \colorbox{pink}{-0.405} &                        0.125 &         \colorbox{pink}{-0.404} &                         0.134 &                          0.002 &                        0.051 &                            -0.024 &                                 0.024 &          \colorbox{pink}{0.413} &          \colorbox{pink}{0.204} & \colorbox{pink}{-0.397} &       \colorbox{pink}{-0.377} &          \colorbox{pink}{-0.356} \\
             153093\_gt.jpg &          \colorbox{pink}{0.401} &        \colorbox{pink}{0.316} &         \colorbox{pink}{0.316} &             \colorbox{pink}{0.468} &                            0.118 &             \colorbox{pink}{0.45} &          \colorbox{pink}{-0.2} &                        0.016 &         \colorbox{pink}{-0.262} &                         0.136 &                          0.023 &       \colorbox{pink}{0.175} &                            -0.077 &                                 0.077 &           \colorbox{pink}{0.31} &                           0.041 & \colorbox{pink}{-0.545} &       \colorbox{pink}{-0.524} & \textbf{\colorbox{pink}{-0.514}} \\
             187029\_gt.jpg &          \colorbox{pink}{0.371} &        \colorbox{pink}{0.228} &         \colorbox{pink}{0.228} &             \colorbox{pink}{0.352} &           \colorbox{pink}{0.263} &            \colorbox{pink}{0.371} &        \colorbox{pink}{-0.201} &                        0.074 &         \colorbox{pink}{-0.226} &                         0.076 &                          0.011 &       \colorbox{pink}{0.361} &            \colorbox{pink}{0.221} &               \colorbox{pink}{-0.221} & \textbf{\colorbox{pink}{0.385}} &                           0.076 & \colorbox{pink}{-0.357} &       \colorbox{pink}{-0.361} &          \colorbox{pink}{-0.367} \\
             198023\_gt.jpg &          \colorbox{pink}{0.372} &        \colorbox{pink}{0.242} &         \colorbox{pink}{0.242} &             \colorbox{pink}{0.458} &                            0.001 &             \colorbox{pink}{0.48} &        \colorbox{pink}{-0.405} &       \colorbox{pink}{0.298} &         \colorbox{pink}{-0.415} &        \colorbox{pink}{0.303} &                          0.131 &       \colorbox{pink}{0.319} &           \colorbox{pink}{-0.177} &                \colorbox{pink}{0.177} &          \colorbox{pink}{0.436} &                          -0.106 & \colorbox{pink}{-0.494} &       \colorbox{pink}{-0.459} & \textbf{\colorbox{pink}{-0.504}} \\
             239096\_gt.jpg &          \colorbox{pink}{0.427} &        \colorbox{pink}{0.414} &         \colorbox{pink}{0.414} &             \colorbox{pink}{0.389} &           \colorbox{pink}{0.312} &            \colorbox{pink}{0.427} &        \colorbox{pink}{-0.383} &      \colorbox{pink}{-0.205} &          \colorbox{pink}{-0.39} &                        -0.127 &                         -0.132 &                       -0.005 &                             0.034 &                                -0.034 &          \colorbox{pink}{0.344} &                           0.127 & \colorbox{pink}{-0.384} &       \colorbox{pink}{-0.463} & \textbf{\colorbox{pink}{-0.441}} \\
             242078\_gt.jpg &          \colorbox{pink}{0.359} &        \colorbox{pink}{0.299} &         \colorbox{pink}{0.299} &             \colorbox{pink}{0.373} &           \colorbox{pink}{0.239} &            \colorbox{pink}{0.423} &        \colorbox{pink}{-0.313} &      \colorbox{pink}{-0.229} &         \colorbox{pink}{-0.292} &                        -0.141 &                         -0.043 &                        0.014 &                            -0.047 &                                 0.047 &                           0.095 &                           0.076 & \colorbox{pink}{-0.367} &       \colorbox{pink}{-0.552} & \textbf{\colorbox{pink}{-0.527}} \\
             323016\_gt.jpg &          \colorbox{pink}{0.446} &                         0.153 &                          0.153 &    \textbf{\colorbox{pink}{0.521}} &           \colorbox{pink}{0.212} &            \colorbox{pink}{0.492} &        \colorbox{pink}{-0.348} &                        0.138 &         \colorbox{pink}{-0.334} &                         0.107 &                          0.056 &      \colorbox{pink}{-0.211} &           \colorbox{pink}{-0.221} &                \colorbox{pink}{0.221} &          \colorbox{pink}{0.317} &          \colorbox{pink}{0.216} & \colorbox{pink}{-0.332} &       \colorbox{pink}{-0.375} &          \colorbox{pink}{-0.439} \\
             376001\_gt.jpg &          \colorbox{pink}{0.228} &        \colorbox{pink}{0.193} &         \colorbox{pink}{0.193} &             \colorbox{pink}{0.227} &                            0.154 &   \textbf{\colorbox{pink}{0.314}} &        \colorbox{pink}{-0.167} &                       -0.046 &         \colorbox{pink}{-0.174} &                        -0.054 &                          0.056 &                        -0.03 &                            -0.037 &                                 0.037 &          \colorbox{pink}{0.263} &                           0.077 & \colorbox{pink}{-0.279} &       \colorbox{pink}{-0.272} &          \colorbox{pink}{-0.298} \\
\hline
                        All &          \colorbox{pink}{0.334} &        \colorbox{pink}{0.181} &         \colorbox{pink}{0.283} &    \textbf{\colorbox{pink}{0.389}} &           \colorbox{pink}{0.194} &            \colorbox{pink}{0.375} &        \colorbox{pink}{-0.277} &                        0.011 &         \colorbox{pink}{-0.289} &        \colorbox{pink}{0.063} &        \colorbox{pink}{-0.089} &                        0.022 &                            -0.015 &               \colorbox{pink}{-0.049} &          \colorbox{pink}{0.302} &          \colorbox{pink}{0.049} & \colorbox{pink}{-0.315} &       \colorbox{pink}{-0.325} &          \colorbox{pink}{-0.341} \\
\hline
\end{tabular}}
\end{table*}

\subsection{Is the ground-truth the perfect colourisation for its grey-scale prior?}
\begin{figure}[ht]
\includegraphics[width=0.5\textwidth]{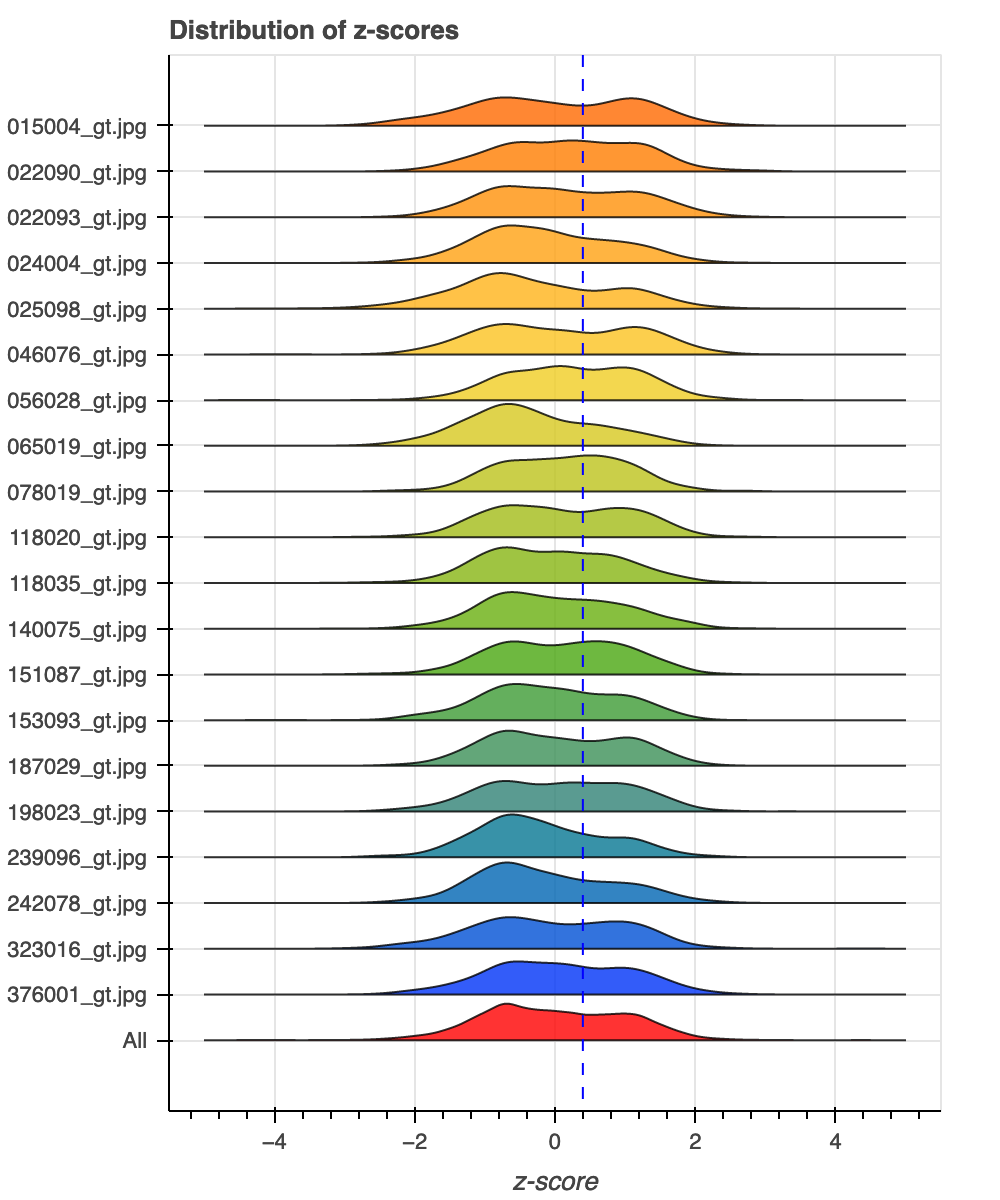}

    \caption{The distribution of responses after the processing in section \ref{ProcRawSec} for all reference images individually and all together. The blue-dashed line shows the value for the ground-truth images. All ground-truth images are assumed equal as we grade on difference scores from the ground-truth. Any area under the curve to the right of the blue line represents scores where the participant gave the recoloured version a higher score than they gave the ground-truth.  We can see that all references have a large area to the right of the ground-truth score. This shows that the ground-truth is far from the most plausible colourisation, as judged by human evaluation.}
    \label{OverallDistro}
\end{figure}

The method currently employed in most deep-learning colourisation systems is to take any natural image dataset, convert the images to CIEL*a*b*, then use the L*-channel as the prior (input) and predict the a*b* colour channels, with the colour channels from the dataset as the ground-truth.
Figure   \ref{OverallDistro} shows that human observers do not rate the ground-truth as higher than all other colour versions that were created in the dataset. Approximately 36\% of the area is above the mean ground-truth score (to the right of the blue-dashed line). This shows that many more plausible colourisations of a scene exist than the ground-truth, but will, in current training regimes, be penalised for being different from the ground-truth.\\
The 20 ground-truth images in our dataset are quite good as they come from the BSD dataset. The images are not necessarily natural or high-quality in many commonly used large image datasets; They may be in black and white, duo-tone, or stylised. 
In classification models, these unnatural or poor quality images are a feature rather than a bug as the desire is to train models to recognise objects even in poor quality images. It therefore makes sense for poor quality images to have the same label as high-quality images if they contain the same object. For generative tasks, such as colourisation, when the task requires a model to generate high-quality colourisations, then poor colourisations in the dataset should not have an equal label to good ones. However, the lack of a reliable no-reference measure for the quality of a colourisation leaves little choice but to treat all images in a dataset as equal-maximum colourisation quality. The only alternative is to assess and sort the large training datasets by resource-intensive human visual inspection.

\subsection{Does white-balance correction of the ground-truth image lead to higher opinion score?}
Photoshop's \cite{Photoshop} white-balance auto-correction was used to produce a white-balance corrected version of each ground-truth image. Using only the direct comparisons between the ground-truth images and WB corrected images resulted in a score of 0.376 for the white-balanced images and a mean of 0.364 for the ground-truth images. This difference is minimal and has a statistical significance of p=0.058, using the Mann-Whitney u-test \cite{MannWhitney10.1214/aoms/1177730491}. While the traditional threshold of $p<0.05$ is arbitrary, the mean difference has not reached that threshold, and with such a slight difference in the value of the mean, white-balance correcting the images in a dataset before training will have only a minimal effect unless there is reason to believe the images in the dataset have particularly bad colour casts.

\begin{figure}[ht]

	\includegraphics[width=0.5\textwidth]{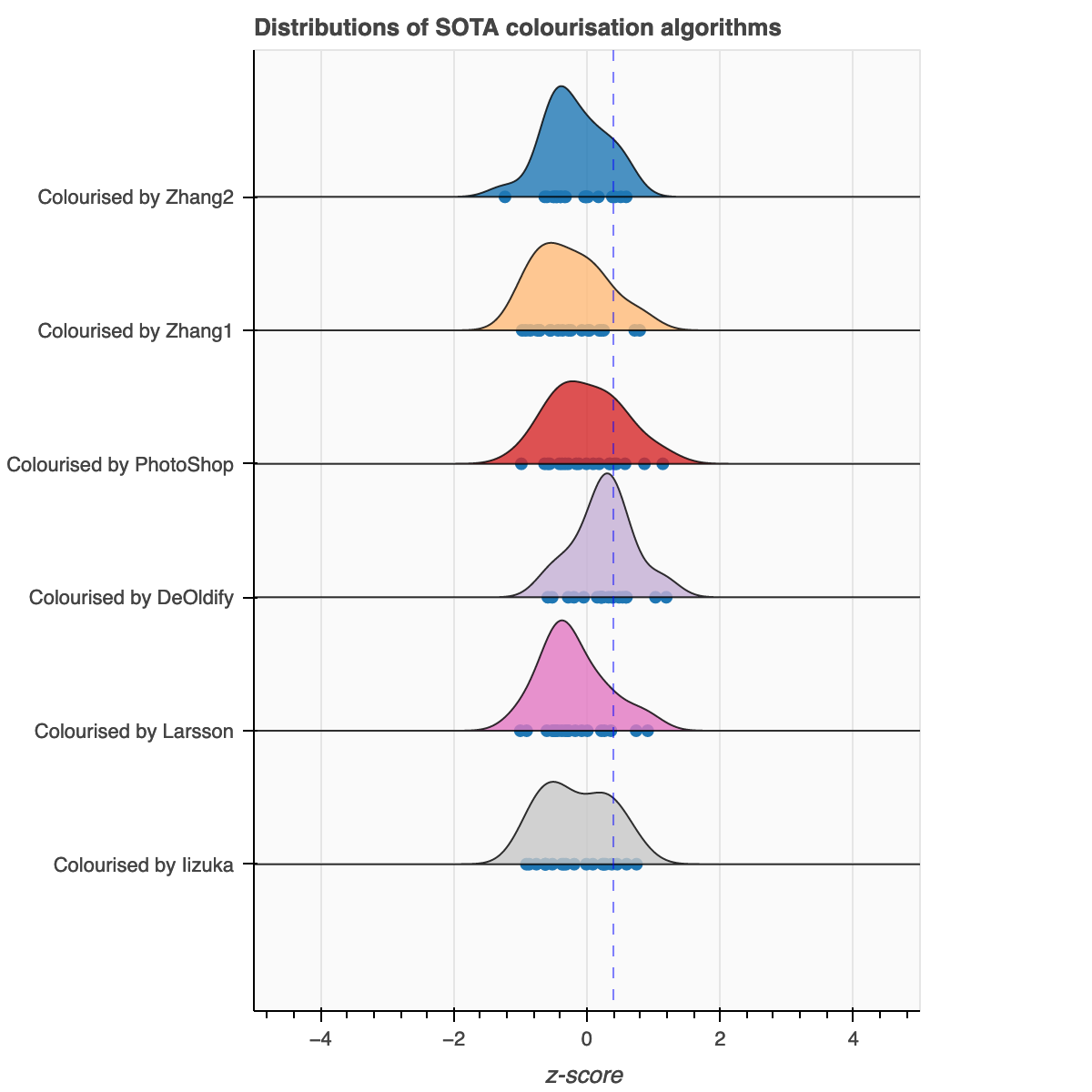}
	\caption{The blue dashed line shows the average Ground-Truth score, when compared against SOTA colourisations, of 0.397. This is higher than the average of the SOTA Colourisation methods, but we can see that all methods achieve some part of the distribution of their scores which is higher than the average ground-truth score. The blue dots represent the mean score of individual images. These can be explored further with our interactive tool.}
	\label{SotaDistros}
 \end{figure}

\begin{table}
\centering
\caption{Mean value Z-Score for the six SOTA methods that we tested, shown in descending order. The mean of the ground-truth when compared with the SOTA algorithms was 0.397.}
\label{SOTAMeans}
\begin{tabular}{lr}
\hline
SOTA Colourisation &  Mean z-score \\
\hline
          Ground-Truth &    0.397\\
          DeOldify &         0.258 \\
         PhotoShop &         0.001 \\
            Iizuka &        -0.151 \\
           Larsson &        -0.196 \\
           Zhang 2 &        -0.201 \\
           Zhang 1 &        -0.264 \\
           
\hline
\end{tabular}
\end{table}

\subsection{How do SOTA colourisation algorithms fair?}
Six state-of-the-art colourisation algorithm's outputs were included in the HECD dataset. The choice of algorithms was made primarily on the availability of implementation and the ability to accept the exact image dimensions used in BSD images. The results in Figure  \ref{SotaDistros} and table \ref{SOTAMeans} show that the two commercial products, DeOldify (from MyHeritage.com \cite{MyHeritage}) and Photoshop \cite{Photoshop}, edge ahead of all the others, which are considerably less recent than the commercial products. DeOldify came top in the surveys, and the difference to Photoshop was statistically significant with a p-value of 0.001, using the Mann-Whitney u-test \cite{MannWhitney10.1214/aoms/1177730491}. The mean score for the ground-truth images was still higher than the mean for any of the SOTA methods, table \ref{SOTAMeans}. \\
Many of the human-evaluation methods outlined in section \ref{LitReview} found that their method could fool a human evaluator or obtain a higher score from a human evaluator on some occasions. We also find this to be the case, as evidenced by the area under the curves in Figure  \ref{SotaDistros} to the right of the dashed line. This area represents the proportion of samples from each model that achieved a higher score than the ground-truth when it and the ground-truth appeared together for comparison scoring.

\begin{figure*}
    
        \includegraphics[width=14.7cm]{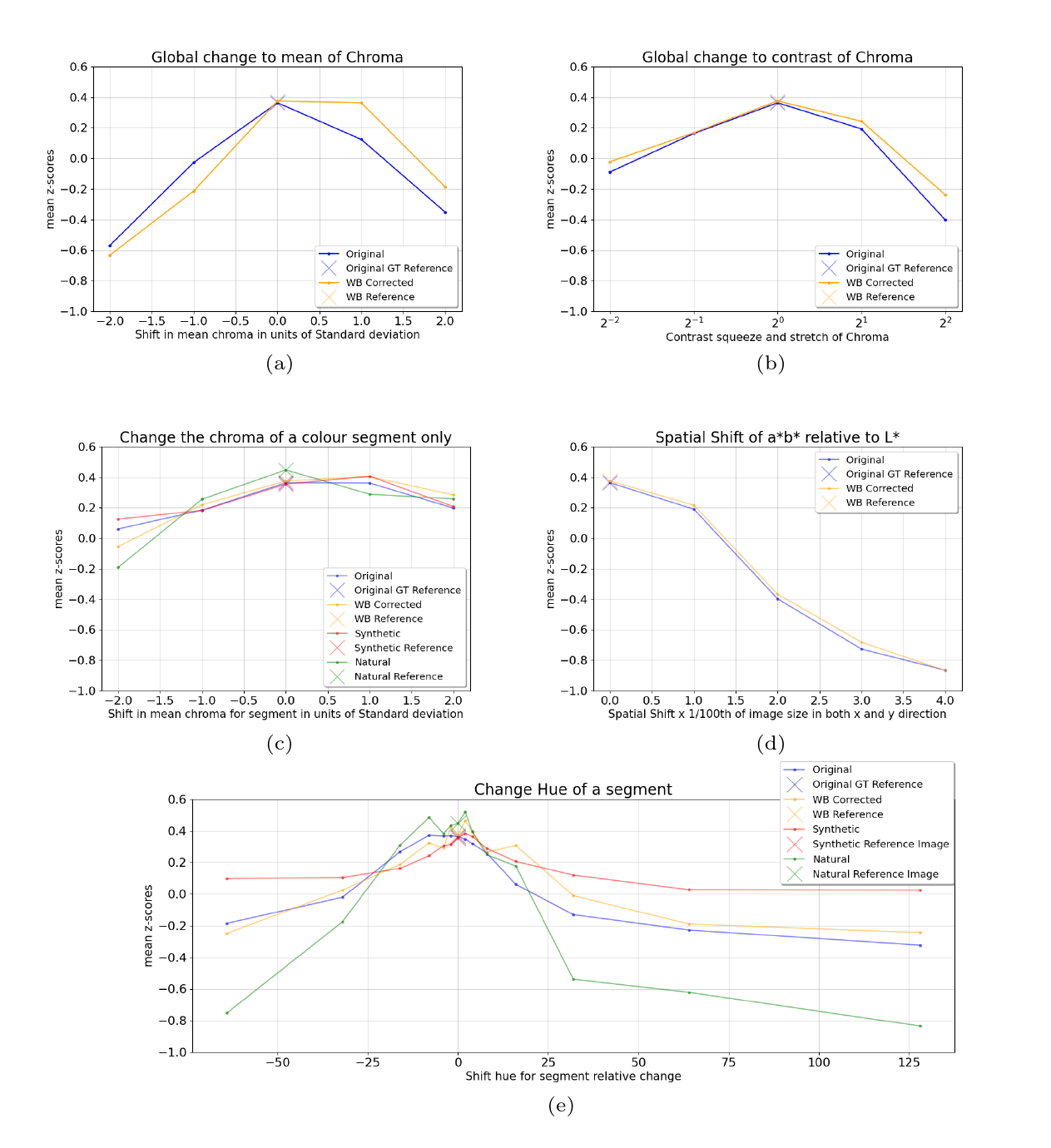}
    
    \caption{In the figure, are various subsets of the data relating to specific modifications, outlined in section \ref{datasetSec}, and the effect of those on mean opinion score. All graphs have the same y-scale (mean opinion z-score) so that comparisons of different types of change can be made at a glance. Sub-figures (a) and (b) look at relative global changes to the statistics of the chroma of the reference image. Sub-figure (c) is the equivalent of (a) but for changes to only a colour segment of an image, leaving all other pixels the same as the reference. (d) shows the effect of spatially shifting (misaligning) the colour channels relative to the L*-channel. (e) looks at the effect of changing the relative hue of a colour segment while leaving all other pixels unchanged from the reference. We look at two data slices for global changes, whether the recolour version derived from the original ground-truth or the white-balance corrected image. For segment modifications (c) and (e), we also look at those slices as well as slicing on whether the modified segment represented a natural or synthetic object.}
        \label{ModResults}
 \end{figure*}

\subsection{Image Modification Statistics}

Figure  \ref{ModResults} shows the effect of making a single modification type to a reference image. These are all relative changes that should be understood in terms of the reference image. The L*-channel is held fixed for all these modifications. Units, such as standard deviation, refer to the statistics of the reference image and so will represent a different absolute value in each case. Figures \ref{ModResults} (a) and \ref{ModResults} (b) show that when the statistics of the chroma of a reference image change, this will, in general, cause a deterioration in mean opinion score, with the caveat that the participants seemed to prefer slightly higher chroma than the reference.  Figure \ref{ModResults} (c) is the equivalent change to \ref{ModResults} (a) but for only a colour segment of the image, with results are similar but less pronounced, as the rest of the pixels in the image retain the reference statistics. 
Figure \ref{ModResults} (d) shows the effect of spatially shifting the colour channels relative to the L*-channel, causing deterioration with an increase in spatial misalignment. However, it should be noted that slight misalignment leads to a relatively small drop in opinion score, particularly when we consider that all pixels are misaligned. We can extrapolate that local colour bleeding across boundaries in colourisations by a small number of pixels will have a relatively small impact on opinion score. Indeed chroma subsampling, widely used in image and video encoding, utilises the Human Visual System's lower acuity in chroma.  
\ref{ModResults} (e) shows the effect of changing the hue of a segment. When the data is split into it's two reference image categories, namely original ground-truth and white-balance corrected image derived from the ground truth, the responses of these are broadly similar. However, when the data is separated into hue changes to colour segments representing natural objects and those representing synthetic objects, a clear difference between the two groups emerges. Examples of natural objects are skin tones and foliage. Examples of synthetic objects are painted surfaces and textiles. Figure \ref{ModResults} (e) shows that both categories see a deterioration in opinion score with medium to large changes in hue for a segment. However, this deterioration is relatively small for synthetic segments compared to the large change  for natural objects. While synthetic objects can theoretically take on any hue, there is still a drop in opinion score with large changes in hue for a colour segment. This may be because the L*-channel prior and the surrounding colours (which did not change from the reference) constrain the most plausible hues to a small band of hue values close to the ground-truth. For colour segments of natural objects, the response is quite different. Small changes in hue to a natural segment may increase the mean opinion score. This may be that the small correction looks more plausible, but it could also be the inherent noise in opinion scores, particularly due to the more dense sampling close to the reference hue. However, the trend is that medium to large changes in natural segment hue sees a large deterioration in the mean opinion score. By directly comparing Figure \ref{ModResults} (e) with Figures \ref{ModResults} (a) and \ref{ModResults} (d) we can see that changing the hue of a natural segment by 64/256 of the full-scale has an equivalent effect on the opinion score of misaligning the colour channels with L* of 0.03 of the dimensions of the image and it has a greater effect than globally changing all of the chroma values by two standard deviations of the chroma in the reference image. This tells us that not all pixels are created equal in colourisation performance.

\section{Conclusion.}

We have shown that the widely-used objective measures utilised in the colourisation literature do not correlate well with human opinion. MS-SSIM shows the highest correlation in our findings but is still too low to make it an appropriate gauge of colourisation quality.\\
The hue of natural objects stands out as significant to the human opinion of the naturalness of an image. Observers seem tolerant of minor differences in hue to natural objects, but medium to large changes are heavily penalised. The observers are relatively tolerant of all changes to the hue of synthetic objects. \\
There is a general trend towards a preference for more saturated (higher average chroma) images. Small increases to the chroma of the ground-truth images led to higher opinion scores, but increases beyond that led to a deterioration in opinion score, as did any decrease in the chroma from the ground-truth. The trends were similar when changes were only made to the chroma of small colour segments; The effects were smaller because only some pixels were affected by the change. However, the effect is not necessarily proportional to the number of pixels, as the observer may be guided by the discrepancy in chroma to the surrounding regions. Both increasing and decreasing the global contrast of chroma caused a deterioration in opinion scores.\\
The observers registered a slight change in opinion for small global registration discrepancies between the colour channels and the L*-channel. Increasing deregistration led to a significant deterioration in opinion scores. This suggests some tolerance to small amounts of colour bleeding but intolerance to more significant amounts. We can assume some cross over with the hue of natural objects here; If de-registration problems change the hue of a natural object, we will again see a significant deterioration in opinion score.\\
Finally, caution should be exercised in simply treating all colour images in a data set as perfect colourisations. The results show that many versions in our limited set of arbitrary modifications scored higher than the ground-truth. Auto-white-balance correction of ground-truth images brought only a minor improvement on average, though it may bring a more significant improvement if the white-balance is poor in the ground-truth images.

\bibliographystyle{apalike}

\bibliography{refs.bib}


\end{document}